\documentclass[twocolumn]{article}

\usepackage{microtype}
\usepackage{graphicx}
\usepackage{subfigure}
\usepackage{booktabs}
\usepackage[]{hyperref}

\usepackage{amsmath}
\usepackage{amssymb}
\usepackage{mathtools}
\usepackage{amsthm}
\usepackage[capitalize,noabbrev]{cleveref}

\usepackage{poker}
\setkeys{poker}{inline=symbol, colorset=4c}
\usepackage{xcolor}

\author{Yifan Yanggong, Hao Pan, Lei Wang}

\title{Mastering the Game of Guandan with Deep Reinforcement Learning and Behavior Regulating}

\begin{document}
\maketitle

\begin{abstract}
    Games are a simplified model of reality and often serve as a favored platform for Artificial Intelligence (AI) research. 
    Much of the research is concerned with game-playing agents and their decision making processes. 
    The game of Guandan (literally, ``throwing eggs'') is a challenging game where 
    even professional human players struggle to make the right decision at times. In this paper we propose a framework
    named GuanZero for AI agents to master this game using Monte-Carlo methods and deep neural networks. The main contribution
    of this paper is about regulating agents' behavior through a carefully designed neural network encoding scheme. 
    We then demonstrate the effectiveness of the proposed framework by comparing it with state-of-the-art approaches.
\end{abstract}

\section{Introduction}
    Games are often considered as an excellent platform for AI research because they not only are a simplified representation of
    the real world, but also have gigantic state space and high complexity. Over the recent years we have
    witnessed an exciting wave of game-playing AI systems being developed. For perfect information games, AlphaGo \cite{silver16} is a 
    significant milestone as it became the first computer program to defeat a professional human 
    Go player on a full-sized board. Later AlphaZero \cite{silver18} demonstrated the feasibility of applying AlphaGo-like 
    algorithms on a broader range of board games such as shogi and chess. Using a similar approach of AlphaZero, 
    MuZero \cite{schrittwieser20} not only achieved superhuman strength, but also surpassed AlphaZero in the mastery of various board games.
    MuZero opened a new horizon where the agent learns the game without even knowing the rules. This breakthrough allowed agents 
    to handle domains with much more complex inputs, such as video games. Many video games incorporate imperfect information, 
    which poses a formidable challenge to AI-related research. Counter-factual regret minimization (CFR) \cite{zinkevich08} was 
    introduced to solve large games with incomplete information by finding an approximate Nash Equilibrium. The authors showed that 
    abstractions of limit Texas Hold'em with as many as $10^{12}$ states can be solved. Also targeting large games, 
    Neural Fictitious Self-Play (NFSP) \cite{heinrich16} was proposed to learn approximate Nash Equilibria by combining fictitious 
    self-play with deep reinforcement learning. NFSP was applied to Limit Texas Hold'em and achieved a win rate (WR) similar to that 
    of the top computer agents. The poker AI DeepStack \cite{moravcik17} used a novel search algorithm 
    where CFR is utilized to update strategy in its lookahead tree while neural networks are used for leaf evaluation. Concurrently 
    with DeepStack, another group of poker AI researchers developed Libratus \cite{brown18} based more heavily on game-theoretic approaches. 
    More recently, using an oracle guiding system to aid the learning process, the mahjong AI Suphx \cite{li20} was able to achieve a higher rating
    than 99.99\% of the human players could manage on Tenhou, one of the most popular online mahjong services in Japan.
    
    Dealing with imperfect information is a challenge already, handling the vast 
    state space in complex games is another. Nonetheless AI agents still achieved impressive results in StarCraft II \cite{vinyals19}, 
    DotA \cite{berner19}, and Honor of Kings \cite{ye20}. However, we should treat these results with a grain of salt. Computer scientist 
    David Churchill stated that ``StarCraft is nowhere near being `solved', and AlphaStar is not yet even close to playing at a 
    world champion level'' \cite{garisto19}. This was consistent with the observations made by many StarCraft II players as they 
    noticed that the AI agent oftentimes did not make the most efficient strategic decisions.
    
    Being much less complex than StarCraft, the poker game Guandan\footnote{https://en.wikipedia.org/wiki/Guandan} enjoys a massive player base. 
    It is estimated that there are at least 130 million Guandan enthusiasts in China. Guandan offers multiple layers of challenge 
    such as imperfect information and its enormous state space. Conventional algorithms such as 
    CFR \cite{zinkevich08} would require extra adjustment when applied to the multi-player setting, especially when one would like to 
    encourage cooperative behavior among the teammates. Even with innovative reinforcement learning (RL) approaches applied to 
    Dou dizhu \cite{jiang19, zha21}, a game similar to Guandan, we noticed that there was a lack of cooperation among the agents, as they were 
    trained to solely maximize their own WR. Additionally, both the number of information sets ($~10^{36}$) and number of 
    legal actions ($~10^{4}$) \cite{lu22} are immense, which could potentially lessen the efficacy of the existing algorithms. For example, 
    action elimination \cite{mnih15} was to be paired with Deep Q-Network(DQN) to eliminate the risk of overestimating errors in the 
    Q-function. Such errors were known to cause the learning algorithm to converge to a suboptimal policy, especially where the 
    action space is huge \cite{thrun93}. It was also found that many RL approaches lacked the ability to generalize 
    over the set of actions \cite{dulac15}. This struggle was confirmed in experiments \cite{you19} where both DQN and A3C \cite{mnih16} agents 
    were applied to the poker game of Dou dizhu. Using DQN, the off-policy target became unstable and overly-optimistic, 
    while A3C saw limited success as it struggled to learn the combinatorial structure of actions. 
    
    Few previous attempts were made to build a Guandan AI agent. Upper Confidence bound applied to Trees (UCT) \cite{kocsis06} was utilized
    to search the game tree based on sampled game states \cite{shen20}. WR and cumulative promotion were used to measure the 
    performance of the Guandan AI agent. The WR eventually rose to 65\% vs random agents whose actions were chosen from all legal actions 
    at random. Another family of Guandan AI Agents \cite{lu22,zhao23,ge24} were developed utilizing the basis established in DouZero 
    \cite{zha21} by incorporating Deep Monte-Carlo (DMC) into an actor-learner scheme to continuously update the Q-values. 
    Although these agents were able to push the WR even higher, one issue remains: there lacks a way to cultivate desired behaviors 
    such as cooperation, or at least to provide a chance for the agents to learn such behaviors.

    In this paper we propose a reinforcement learning framework GuanZero for AI agents to not only master the game of Guandan, 
    but also gain an understanding of the desired behavior in an efficient way. The proposed framework 
    will be relying on DMC to utilize its great scalability \cite{zha21}, while at the same time 
    cultivating cooperative and other desired behavior through a carefully designed neural network encoding scheme.

\section{Background of Guandan}
    Guandan is a four-player, fixed-partnership shedding-type climbing card game originated in the Jiangsu province of China. 
    In recent years, Guandan quickly rose to national fame and made its way to the 5$^{th}$ National Mind Sports Games. 
    Two standard 
    52-card decks plus two red jokers and two black jokers are used. Each player starts with 27 cards. One of the unique traits of 
    the game is the numerous playable combinations (detailed in \cref{cardcombos}), making the game quite entertaining. A
    powerful card combination allowed in the game is called a bomb, which is a homophone of the word ``egg'' in Chinese.
    Literally, the game of Guandan is about playing bombs, which is difficult to master, 
    as a player possesses no more than three bombs on average.

    Guandan is related to another poker game called Dou dizhu but distinguishes from it due to its level card system. 
    Each team starts with their own
    level card set to two and race against the other team to upgrade their level card to ace. 
    The level card not only serves as a scoring system for 
    the two teams, but also offers powerful perks. For example, the current level card trumps everything except the joker cards, 
    and both level cards of hearts 
    become the wild cards. Wild cards cannot represent either of the joker cards but everything else, making them quite powerful utility wise.
    For example, one can create powerful card combinations such as a bomb or even a straight flush using the wild cards, which would
    be impossible otherwise due to the player lacking specific card(s) for the desired combination.
    
    A full Guandan game consists of several mini games. We define a mini game to be the series of all actions taken to decide the standing
    of each player for a fixed level card. A team wins a mini game when one of its members becomes the first to empty their hand cards 
    before all three other players. This player is called the banker. Once a mini game concludes, 
    the winning team's level card can be upgraded by up to three ranks depending on the standing of the team members. The whole game ends when 
    the winning team's level card is Ace and none of the team members on the winning team finishes the last. 
    For more detailed cardplay rules please refer to \cref{rules}

    At the start of each mini game (except the very first one), players must carry out a tribute process where the last finisher (dweller) 
    of the previous mini game donates his highest-ranked card to the banker. Wild cards are not counted as the highest-ranked 
    cards. To balance out the number of cards each player has, the banker must return a card with the rank no higher 
    than ten. Since the tribute process is separate from the actual cardplay, we did not apply reinforcement learning to help the agents decide 
    which card to donate/return, but rather employed a rule-based approach which suffices most of the time. 
    More details of the tribute process can be found in \cref{tribute}.

\section{Methodology}
    In this section we describe the building blocks of our proposed reinforcement learning framework for Guandan. Reinforcement learning is a process
    for the agents to learn how to map situations to actions and the resulting reward \cite{sutton18}. 
    One frequently-encountered hurdle here is the evaluation of the value pertaining to a specific action. This becomes more evident 
    for complex games. It was even stated that, 
    ``no simple yet reasonable evaluation function will ever be found for Go'' \cite{mueller02}. One cannot solve this problem 
    using brute force since the vast
    state space would make any exhaustive search nearly impossible to complete. Fortunately, we can model Guandan as a series of episodic tasks to
    harness their property of ensuring well-defined returns. Monte-Carlo methods are designed for such tasks and they will be introduced next.

\subsection{Monte-Carlo Methods and Deep Neural Networks}
    Monte-Carlo (MC) methods are simple yet effective ways to estimate the value function and in turn help discovering the optimal policies. 
    They are simple because no complete knowledge of the environment is required except the experience which includes states, actions,
    and rewards acquired through the interaction with the environment. Such experience can even be acquired through simulations. MC methods solve
    reinforcement learning by averaging sample returns. Effectively, this is to estimate the expected return $q_{\pi}(s, a)$ when starting
    in state $s$, taking action $a$, and following policy $\pi$. However, prior to this, we must generate an episode to obtain the aforementioned
    $s$ and $a$ using $\pi$, and this is tricky to do. Due to the scale of the state space, many state-action pairs may never be visited.
    This situation becomes worse if $\pi$ is deterministic. Following $\pi$, one can only observe returns for one action from each state. 
    MC methods cannot progress properly because there is no returns to average. In turn, experience cannot be improved through learning. 
    Instead, we must consider all the actions from each state, not just the one(s) favored by the current policy $\pi$. This is the general
    problem of balancing exploration and exploitation, as in the context of the k-armed bandit problem \cite{auer02,katehakis87}. 
    Epsilon-greedy \cite{sutton18} is a simple yet effective approach to balance the two by choosing exploitation most of the time with a small chance of
    exploring. Even with good balance of exploration and exploitation, the estimation of the expected return $q_{\pi}(s, a)$ is still difficult.
    Naturally, deep neural networks can be used for such a task. Combining deep neural networks and MC methods, DMC already saw its success 
    in the game of Dou dizhu \cite{zha21}. Additionally, parallelization of DMC is done easily, making it possible to generate numerous samples
    quickly and in turn alleviate the issue of MC methods having high variance \cite{sutton18}. In the next section we discuss how we encode 
    states and actions which are to be fed into the neural network as inputs.

\subsection{State Representation}
\begin{figure}[ht]
        \vskip 0.2in
        \begin{center}
        \centerline{\includegraphics[width=\columnwidth]{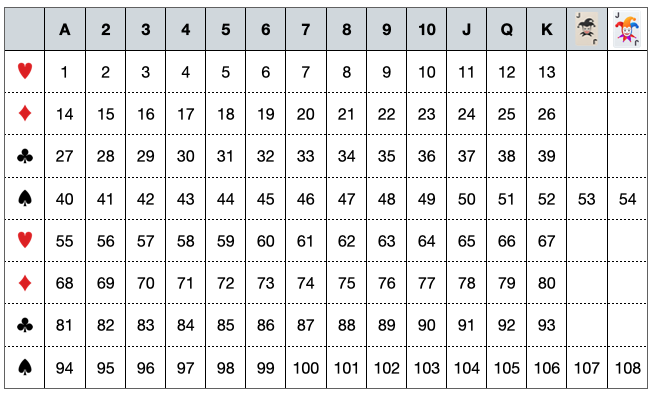}}
        \caption{Indexing scheme of cards}
        \label{fig-index_scheme}
        \end{center}
        \vskip -0.2in
    \end{figure}

\begin{figure}[ht]
        \vskip 0.2in
        \begin{center}
        \centerline{\includegraphics[width=\columnwidth]{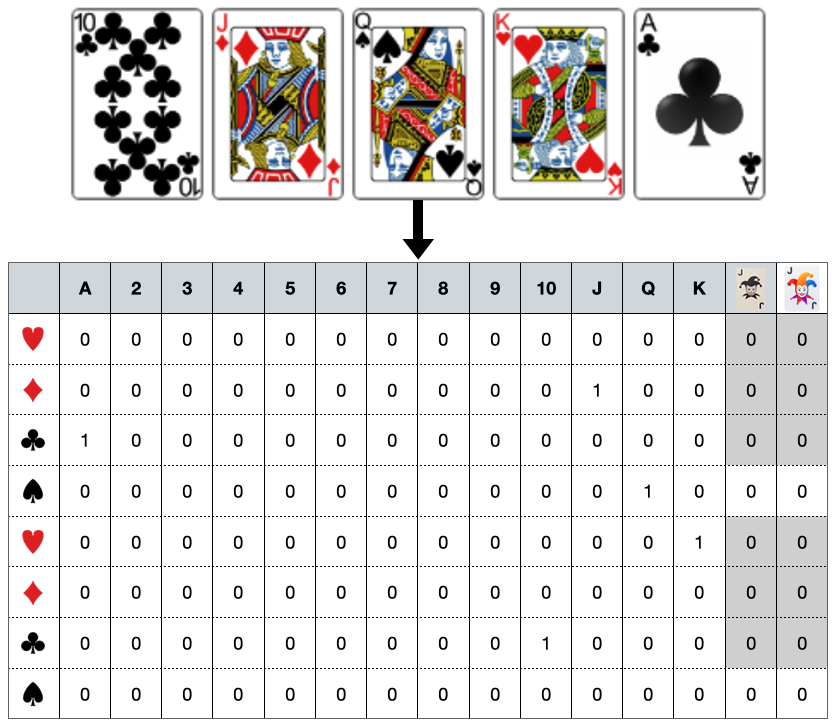}}
        \caption{State representation of cards}
        \label{fig-encoding}
        \end{center}
        \vskip -0.2in
    \end{figure}

    There are a total of 108 cards in the game of Guandan. The card suits matter since the straight flush combination is legal.
    What's more, it is important to keep track of how many cards of a particular suit and rank are remaining. Based on these rationals, 
    we treat each card as a unique entity and designate a number in the interval of $[1, 108]$ to it. One can visualize such an indexing scheme using an 
    $8\times15$ matrix as shown in \cref{fig-index_scheme}. In this matrix, each row is dedicated to a specific suit, with the exception of
    the fourth and eighth rows which also include the four joker cards. An example matrix to encode a specific combination of cards
    is illustrated in \cref{fig-encoding}. The entries corresponding to the existing cards in the combination 
    are set to ones while all others zeros.
    Before being used as an input to the neural network, each matrix is to be flattened, excluding the twelve entries
    which are always zeros (the greyed-out cells in \cref{fig-encoding}). The flattening of the matrix results in a one-hot vector of size 108. 

    The matrix is used to represent several features deemed necessary to depict the state. Such features include but are not limited to: 
    a player's hand cards, the union of hand cards of the other three players, and played cards of each other player. 
    It should be noted that 
    we also record the most recent 20 actions to mimic the reality: a human player often considers what was done recently before 
    making their own decisions. All these features alongside other special ones dedicated to regulate agents' behavior 
    are summarized in \cref{tab-states_actions}. 
    In the next section we will be discussing the behaviors and how we have the agents to learn them via the special one-hot vectors.

    \begin{table*}[t]
        \caption{Features included in the state representation, and action}
        \label{tab-states_actions}
        \vskip 0.15in
        \begin{center}
        \begin{small}
        \begin{sc}
        \begin{tabular}{lll}
        \toprule
        & description & size \\
        \midrule
        \midrule
        State & current player's own hand cards & 108 \\
            & hand cards of all other players & 108 \\
            & most recent action of each player & $108\times4$ \\
            & played cards of each other player & $108\times3$ \\
            & number of remaining cards of each other player & $27\times3$ \\
            & current level card & 13 \\
            & most recent 20 actions & $108\times4\times5$ \\
            & status of cooperating & 3 \\
            & status of dwarfing & 3 \\
            & status of assisting & 3 \\
        \midrule
        Action & the card combination chosen by the current player & 108 \\
        \bottomrule
        \end{tabular}
        \end{sc}
        \end{small}
        \end{center}
        \vskip -0.1in
    \end{table*}

    \subsubsection{Cooperating}
    We define the cooperating behavior in the context of Guandan as the act of a player choosing to pass when they have any legal 
    card combinations that can trump their teammate's which was played recently. The word ``recently'' refers to the most recent two actions.
    The cooperating behavior is often preferred by human players because the goal of Guandan is to empty one's hand cards as fast as possible, 
    and helping the teammate to do so is equally important.
    When the cooperating behavior is indeed chosen, the teammate would likely obtain the right to play subsequently, which is a huge advantage in the game.
    Players often fight for such an advantage using powerful card combinations such as jokers or even bombs. 
    However, the cooperating behavior is sometimes not conducted. The reasons can be various such as the teammate having a really bad hand 
    or perhaps the acting player being much closer to the victory than the teammate.

    There are several conditions to be met before one can execute the cooperating behavior:
    \begin{itemize}
        \item The current player's teammate played a card combination recently.
        \item The opponent following the teammate did not trump the card combination.
        \item The current player has at least one legal card combination which trumps their teammate's.
    \end{itemize}

    The corresponding one-hot vector is designed in such a way to serve as an indication to flag the status of cooperating. 
    The vector is of length three.
    When the conditions of performing a cooperating behavior is not met, we set the vector as $[1, 0, 0]$. When all of the conditions are met, 
    there are only two scenarios to consider: the player either chooses to perform the cooperating behavior, or refuses to do so. The resulting 
    one-hot vectors are $[0, 1, 0]$ and $[0, 0, 1]$, respectively. All the one-hot vectors relating to the cooperating behavior 
    are listed in \cref{tab-cooperating}. Here our design principle is to provide a simple yet logically tight 
    mechanism for the agents to learn when to cooperate and when not to.

    \begin{table}[t]
        \caption{The status of cooperating and the corresponding one-hot vector representation}
        \label{tab-cooperating}
        \vskip 0.15in
        \begin{center}
        \begin{small}
        \begin{sc}
        \begin{tabular}{lcr}
        \toprule
        Scenario & Representation \\
        \midrule
        cannot cooperate & $[1, 0, 0]$ \\
        chooses to cooperate & $[0, 1, 0]$ \\
        refuses to cooperate & $[0, 0, 1]$ \\
        \bottomrule
        \end{tabular}
        \end{sc}
        \end{small}
        \end{center}
        \vskip -0.1in
    \end{table}

    To measure how frequently an agent performs cooperating, we define a metric called the cooperating rate to be the number of times the agent
    does indeed cooperate over the number of times the conditions for the agent to cooperate are met. It should be noted that
    the cooperating rate and the WR are not
    necessarily positively correlated, but in certain scenarios cooperating is favored since doing so could potentially bring the teammate 
    one-step closer to the victory.

    \subsubsection{Dwarfing}
    We also consider the behavior of dwarfing which we define as the act of a player choosing to play a card combination whose size is 
    strictly greater than the minimum
    hand size of any of the two opponents, making it exceptionally difficult for the winning opponent to have any answers.
    The conditions to execute the dwarfing behavior are as follows: 
    \begin{itemize}
        \item It is the current player to lead the play.
        \item The current player has at least one legal card combination whose size is strictly greater than the minimum hand size of the two opponents.
    \end{itemize}

    The dwarfing behavior is frequently executed especially when a particular opponent has very few ($\leq3$) cards in hand, for in this situation
    it is impossible to form a bomb type of card combination as a strong answer. 
    Similar to how we treat the cooperating behavior, we design the one-hot vectors to flag the status of executing the dwarfing behavior, and
    show them in \cref{tab-dwarfing}. We also define the dwarfing rate as the number of times the agent dwarves over the number of times the conditions 
    for the dwarfing behavior are met.

    \begin{table}[t]
        \caption{The status of dwarfing and the corresponding one-hot vector representation}
        \label{tab-dwarfing}
        \vskip 0.15in
        \begin{center}
        \begin{small}
        \begin{sc}
        \begin{tabular}{lcr}
        \toprule
        Scenario & Representation \\
        \midrule
        cannot dwarf & $[1, 0, 0]$ \\
        chooses to dwarf & $[0, 1, 0]$ \\
        refuses to dwarf & $[0, 0, 1]$ \\
        \bottomrule
        \end{tabular}
        \end{sc}
        \end{small}
        \end{center}
        \vskip -0.1in
    \end{table}

    \subsubsection{Assisting}
    We treat the behavior of assisting as well. We define the assisting behavior as the act of a player playing a card combination whose size is 
    smaller than the teammate's hand size, making it easier for the teammate to have an answer to the play. The conditions for an player to
    execute the assisting behavior are as follows:
    \begin{itemize}
        \item It is the current player to lead the play.
        \item The current player has at least one legal card combination whose size is less than that of their teammate.
        \item The card combination to be played must not be of the highest rank (e.g., triple level cards).
    \end{itemize}
    
    We design the one-hot vector representation (see \cref{tab-assisting}) and define the assisting rate similar to those detailed previously 
    for the other two behaviors.
    The assisting behavior is quite useful especially when the teammate's hand size is small ($\leq3$), as it can sometimes take only one action 
    (e.g., playing a single card) for the teammate to empty their hand completely.
    The assisting behavior is sometimes compounded with the dwarfing behavior, as chances are that both the teammate and at least one of the
    opponents have a small-sized hand. But once again, we leverage the strong learning capability of the deep neural network to help to decide
    which behavior to perform and/or when to do so. The structure of the neural network will be described in the next section.

    \begin{table}[t]
        \caption{The status of assisting and the corresponding one-hot vector representation}
        \label{tab-assisting}
        \vskip 0.15in
        \begin{center}
        \begin{small}
        \begin{sc}
        \begin{tabular}{lcr}
        \toprule
        Scenario & Representation \\
        \midrule
        cannot assist & $[1, 0, 0]$ \\
        chooses to assist & $[0, 1, 0]$ \\
        refuses to assist & $[0, 0, 1]$ \\
        \bottomrule
        \end{tabular}
        \end{sc}
        \end{small}
        \end{center}
        \vskip -0.1in
    \end{table}

\subsection{Neural Network Architecture}
\begin{figure}[ht]
        \vskip 0.2in
        \begin{center}
        \centerline{\includegraphics[width=\columnwidth]{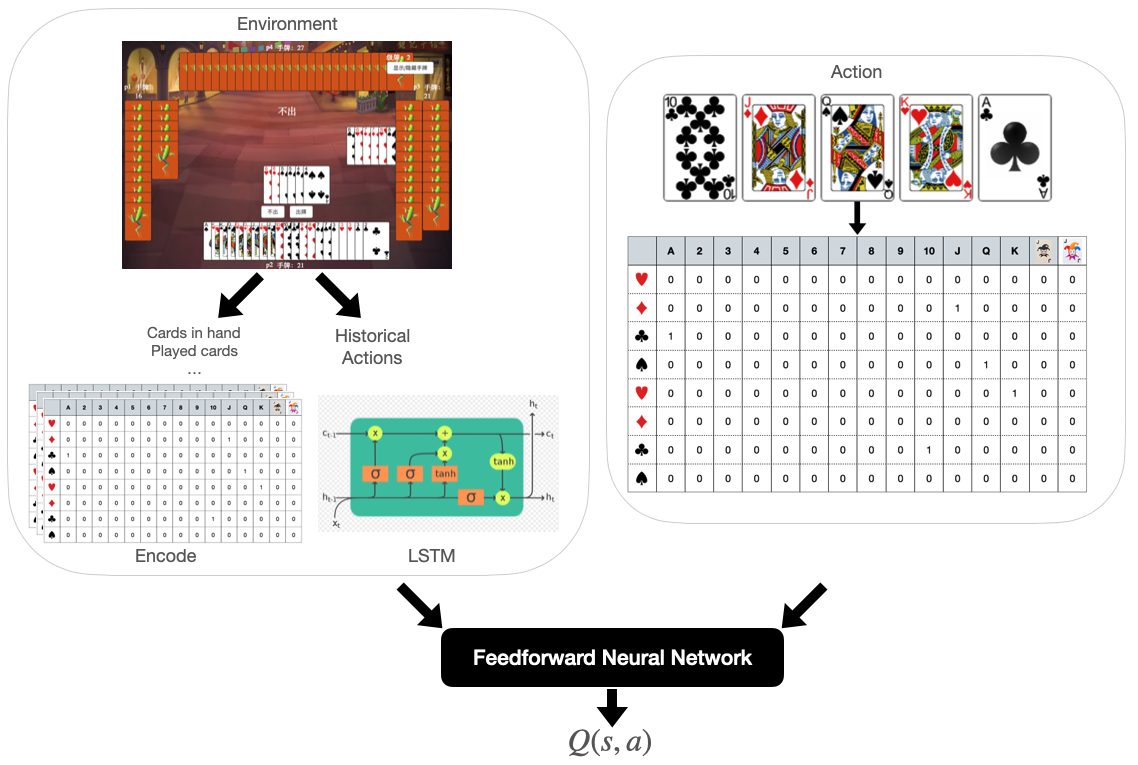}}
        \caption{Network architecture of GuanZero}
        \label{fig-network}
        \end{center}
        \vskip -0.2in
    \end{figure}

    The neural network takes state $s$ and action $a$ as inputs, and estimates the resulting expected cumulative reward $Q(s, a)$. 
    The state is represented by a rich combination of features which are already detailed in \cref{tab-states_actions}. To properly treat the
    historical actions, we employ a Long Short-Term Memory (LSTM) network \cite{hochreiter97} to capture the long-term dependency 
    among action, state, and value. This is achieved by imbuing the network with the ability to learn when to remember and 
    when to forget pertinent 
    information \cite{gers00}. In the mean time, the vanishing gradient problem can be mitigated by LSTM \cite{hochreiter91}, as the 
    network allows the gradients to flow unchanged. Such properties of LSTM in turn facilitates the learning process. 
    With the special treatment of historical actions in place, all the features in 
    the state, together with the action (as depicted in \cref{tab-states_actions}), are concatenated and fed into a 
    feedforward neural network \cite{zell94} consisting of six dense layers with the Rectified Linear Unit (ReLU) \cite{brownlee19} 
    as activation function. The architecture of the neural network is illustrated in \cref{fig-network}.

\subsection{Distributed Learning}
\begin{figure}[ht]
        \vskip 0.2in
        \begin{center}
        \centerline{\includegraphics[width=\columnwidth]{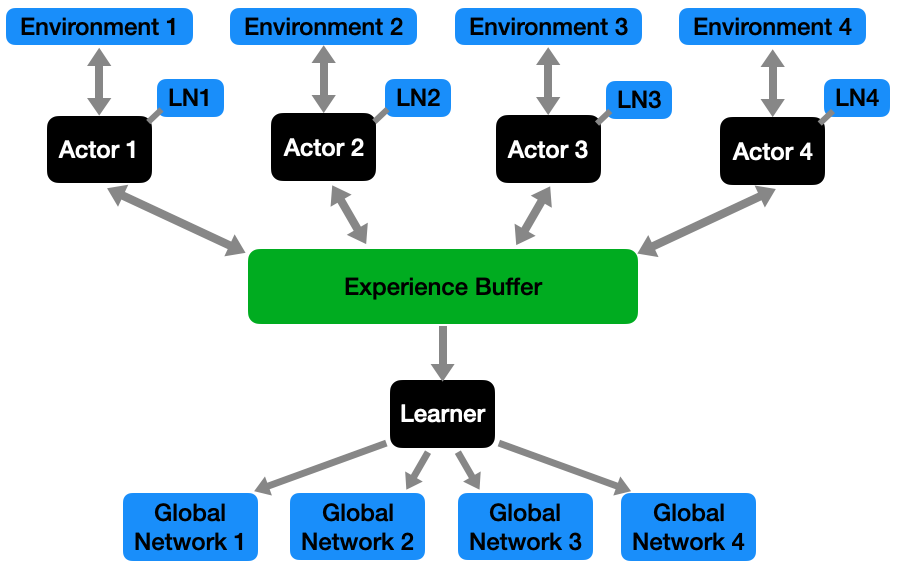}}
        \caption{The distributed learning process of GuanZero}
        \label{fig-actor_learner}
        \end{center}
        \vskip -0.2in
    \end{figure}

    Inspired by A3C \cite{mnih16}, a framework for deep reinforcement learning, we set up our own distributed learning process as follows. 
    First we assign four networks named `p1', `p2', `p3', and `p4' to the four players based on the order of play 
    in the first mini game. We then parallelize the learning process by distributing the simulation tasks to four actors.
    Each of the four actors maintains a local network (LN) when simulating the agent's interaction with their own environment. 
    These local networks
    are synchronized periodically with the four global networks maintained by the learner process. The learner
    process updates those global networks based on the experience gained from the actors. The true Q-values
    is continuously improved by minimizing the Mean Squared Error (MSE). The schematic plot of the distributed learning
    process is depicted in \cref{fig-actor_learner}
    
\section{Experiments}
    In this section we set up experiments to compare the proposed approaches with several alternatives. We also assess the effectiveness of
    behavior regulating on the Guandan agents. When comparing two different types of agents, 
    we form two teams, namely, team A and team B, with the `p1' player of team A to lead the play in the first mini game, followed by
    players `p2', `p3', and `p4'.
    We denote the two players on team A as `p1' and `p3', while those on team B `p2' and `p4'. We restrict the same type of 
    agents to play on the same team. We then run a fixed amount (1000) of mini games using a series of pre-generated 
    random decks. We then swap teams for the four players, and run the same amount of mini games using the same decks again.
    We use WR as the sole metric to evaluate the strength of an agent.

    \subsection{Opposing Guandan Agents}
    We make use of multiple types of Guandan agents to serve as our opponents. 
    The most obvious type of agent is one which picks all actions at random. We call this type of agent
    the random agent. We also pitch our GuanZero agent against a stronger opponent: the champion
    of the first China Guandan AI Algorithm Competition (CGAIAC)\footnote{http://gameai.njupt.edu.cn/gameaicompetition/}.
    This type of agent is rule-based, and we denote it as CGAIAC. 
    However, we believe that non-rule-based agents utilizing reinforcement learning would be even stronger,
    and construct such agents using the framework of DouZero \cite{zha21}, an agent learns to master a similar game, Dou dizhu. 

    As both the DouZero-based and GuanZero agents require training, we train them until convergence is observed. We then pitch
    our GuanZero agents against all the opponents described earlier, and record the WR in \cref{tab-wr_vs3}.

    \begin{table}[t]
        \caption{Win rate of the GuanZero agents vs all opponents, when playing as team A, and as team B, respectively}
        \label{tab-wr_vs3}
        \vskip 0.15in
        \begin{center}
        \begin{small}
        \begin{sc}
        \begin{tabular}{lccr}
        \toprule
        & team A & team B \\
        \midrule
        vs Random & 99\% & 97\% \\
        vs CGAIAC & 82\% & 81\% \\
        vs DouZero-based & 75\% & 77\% \\
        \bottomrule
        \end{tabular}
        \end{sc}
        \end{small}
        \end{center}
        \vskip -0.1in
    \end{table}

\begin{figure}[ht]
        \vskip 0.2in
        \begin{center}
        \centerline{\includegraphics[width=\columnwidth]{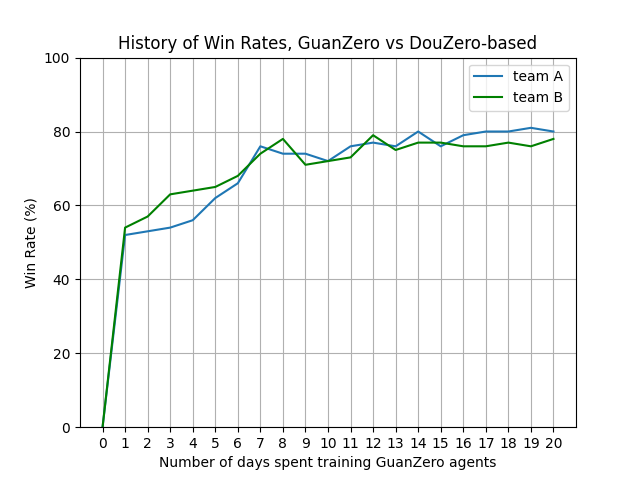}}
        \caption{History of WR achieved by GuanZero agents playing against DouZero-based ones}
        \label{fig-WRvDZ}
        \end{center}
        \vskip -0.2in
    \end{figure}

    First thing we noticed is that GuanZero agents achieved a crushing victory over the random agents, as the random agents do not
    `know' how to make good decisions most of the time. The GuanZero agents faced moderate resistance from the rule-based CGAIAC agents.
    Once a sufficient amount of simulations is reached, RL-based agents start to shine as they become more and more capable of finding
    counter moves against those made by rule-based agents which are constrained by the heuristics/human experience 
    they are relying on. GuanZero agents initially encountered stiff resistance against DouZero-based ones as both agents were trained 
    through self-play and have a deep understanding of the game. Nonetheless, GuanZero agents quickly gained the upper hand
    as the training goes on. The training efficiency is quite satisfactory as it took us less than a week to observe signs of convergence. 
    The WR continued to rise against the DouZero-based agents over the entire course of training.
    (see \cref{fig-WRvDZ}). 

    We contribute GuanZero agents' edge over the DouZero-based ones to the behavior regulating scheme, 
    which we will be analyzing further in the next section.

    \subsection{Behavior Regulating}
    In this section we make use of a few example scenarios to analyze the effect of behavior regulating on our GuanZero agents. 
    For all the examples we discuss in this section,
    we pitch the GuanZero agents against the DouZero-based ones. We believe this is the best setup to show how GuanZero agents
    make rational decisions as the two types of agents are closely matched. 

    \subsubsection{Cooperating}
\begin{figure}[ht]
        \vskip 0.2in
        \begin{center}
        \centerline{\includegraphics[width=\columnwidth]{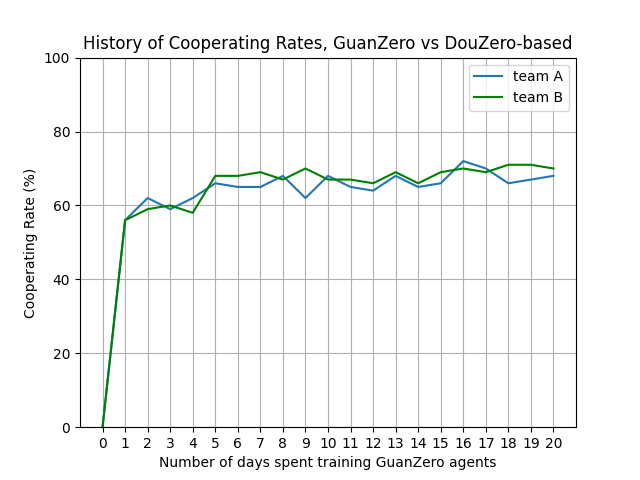}}
        \caption{Evolution of GuanZero agents' coopering rate over the course of training}
        \label{fig-COOPvDZ}
        \end{center}
        \vskip -0.2in
    \end{figure}

    Over the course of training GuanZero agents learn the proper timings to execute the cooperating behavior. 
    The cooperating rate of GuanZero agents stabilizes around a level 
    (shown in \cref{fig-COOPvDZ}) that is significantly higher than the baseline value (roughly 33\%) we observe from random agents.

    GuanZero agents frequently choose the cooperating behavior. We have observed numerous occasions where such decisions
    resulted in GuanZero agents winning the mini game. Here (\cref{tab-coop1}) is a typical scenario 
    of them doing so. GuanZero agents play as team B. 
    Player $p2$ leads by playing \twod, a level card. Following $p3$, $p4$ also chose to pass, 
    despite possessing multiple cards that can trump the \twod played earlier by the teammate $p2$. 
    From the perspective of a human player, this decision is reasonable because
    $p2$'s hand size (8) is much smaller than that (11) of $p4$. This means $p2$ has a much better chance of emptying their hand.
    The decision eventually helped players $p2$ and $p4$ to secure the banker and the follower standings, respectively.

    \begin{table}[t]
        \caption{Hand cards of each player at the time player $p4$ executes the cooperating behavior}
        \label{tab-coop1}
        \vskip 0.15in
        \begin{center}
        \begin{small}
        \begin{sc}
        \begin{tabular}{llr}
        \toprule
        & hand cards \\
        \midrule
        $p1$ & \treh\fives\sixh\sevh\eigh\eigd\ninec\nines\tend\tenc\Qh\Qh\Qc\Kh\Kd\Ac \\
        $p2$ & \twod\sevc\tenh\tend\Js\Qd\Ad jkr {\color{red}Jkr} \\
        $p3$ & \twoc\fives\eigs\Qd\Qc\Qs\Ks\As\Ah\Ad\Ac\Ac \\
        $p4$ & \twos\sixs\ninec\tenc\tenh\Qs\Kh\Kc\Ah jkr {\color{red}Jkr} \\
        \bottomrule
        \end{tabular}
        \end{sc}
        \end{small}
        \end{center}
        \vskip -0.1in
    \end{table}

    \subsubsection{Dwarfing}
\begin{figure}[ht]
        \vskip 0.2in
        \begin{center}
        \centerline{\includegraphics[width=\columnwidth]{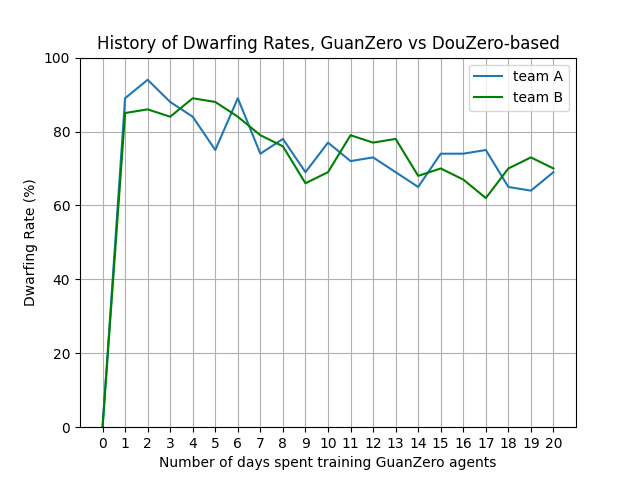}}
        \caption{Evolution of GuanZero agents' dwarfing rate over the course of training}
        \label{fig-DWARFvDZ}
        \end{center}
        \vskip -0.2in
    \end{figure}

    Compared to the other two behaviors, dwarfing happens much less often due to its conditions being more complex. This means
    it is generally more difficult to master the timing of executing this behavior due to the relatively scarce occurrence. Nonetheless,
    GuanZero agents learn when to execute such behavior. As we can see from \cref{fig-DWARFvDZ}, compared to the cooperating behavior, 
    there is much more fluctuation in the dwarfing rate, and it takes the agents much longer to settle around an ideal value.
    
    An instance of executing the dwarfing behavior is demonstrated in \cref{tab-dwar1}. GuanZero agents
    are playing as team A. Player $p3$ leads by playing \eigd\eigh\eigd\nines\nined, dwarfing the winning opponent player $p2$ 
    as the hand size is
    only four. Although GuanZero agents did not end up becoming the winning team, the decision was sound and helped to prevent 
    the more disastrous outcome of the team members receiving the double-dweller status. 

    \begin{table}[t]
        \caption{Hand cards of each player at the time player $p3$ executes the dwarfing behavior}
        \label{tab-dwar1}
        \vskip 0.15in
        \begin{center}
        \begin{small}
        \begin{sc}
        \begin{tabular}{llr}
        \toprule
        & hand cards \\
        \midrule
        $p1$ & \tred\fourc\fives\ninec\tenh\tend\tenc\Jd\Jc\Kd\Kh\Ac \\
        $p2$ & \Jd\Ks\Kd {\color{red}Jkr} \\
        $p3$ & \fourh\eigd\eigh\eigd\nines\nined\Jc\Js\Qd\Qc\Qs\Kc\Ks\Ad\Ah \\
        $p4$ & \eigs\nined\tend\Kc\Ac\ jkr \\
        \bottomrule
        \end{tabular}
        \end{sc}
        \end{small}
        \end{center}
        \vskip -0.1in
    \end{table}

    \subsubsection{Assisting}
\begin{figure}[ht]
        \vskip 0.2in
        \begin{center}
        \centerline{\includegraphics[width=\columnwidth]{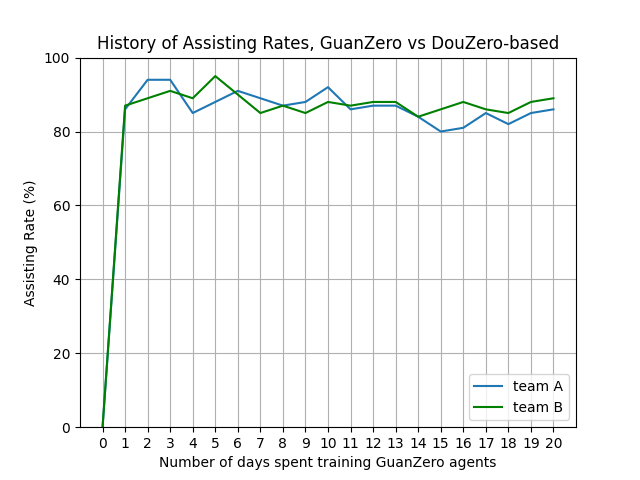}}
        \caption{Evolution of GuanZero agents' assisting rate over the course of training}
        \label{fig-ASSISTvDZ}
        \end{center}
        \vskip -0.2in
    \end{figure}

    Assisting is fairly often executed by the GuanZero agents. 
    We observe similar trends throughout the training (see \cref{fig-ASSISTvDZ}) 
    compared to those associated with the dwarfing behavior.

    A simple example in \cref{tab-assi1} illustrates this behavior. GuanZero agents are playing as team A this time. 
    Player $p3$ leads by playing \twoh, the lowest-ranked single card, 
    utilizing the encoded information that the teammate is on the verge of victory as $p1$ has only 4 hand cards. 
    It should be noted
    that $p3$ did not choose to play certain card combinations (e.g., \tenh\tenc\tens\Qs\Qh) that would empty the hand much faster.
    $p1$ answered effortlessly
    by playing the red joker. Eventually $p1$ and $p3$ managed the banker and the follower, respectively.

    \begin{table}[t]
        \caption{Hand cards of each player at the time player $p3$ executes the assisting behavior}
        \label{tab-assi1}
        \vskip 0.15in
        \begin{center}
        \begin{small}
        \begin{sc}
        \begin{tabular}{llr}
        \toprule
        & hand cards \\
        \midrule
        $p1$ & \trec\fived {\color{red}Jkr} {\color{red}Jkr} \\
        $p2$ & \twod\tres\tred \dots (22 cards) \\
        $p3$ & \twoh\fourd\sevh\nineh\tenh\tenc\tens\Jc\Qs\Qh\Ac jkr \\
        $p4$ & \treh\fivec\sixc\sevd\eigd\eigs\ninec\tenc\Qs\Kd\Ks \\
        \bottomrule
        \end{tabular}
        \end{sc}
        \end{small}
        \end{center}
        \vskip -0.1in
    \end{table}

\section{Conclusion}
    In this paper we propose a deep reinforcement learning framework for agents to master the game of Guandan. We utilize DMC to
    aid the estimation of the cumulative return resulted from taking a specific action in a certain state. 
    The successful operation of DMC relies on simulating a large amount of interaction with the environment. We set up a
    distributed learning process to facilitate the successful operation of DMC. 
    We also regulate the agents' behavior through a carefully-designed
    encoding strategy. Behavior regulating proved to be effective in boosting the learning agents' performance
    in the experiments we conducted. Using these experiments, we were able to demonstrate the superiority of our Guandan agents 
    over all the alternatives we are aware of.

    In the future we are interested in exploring more research areas. The tribute process could use a separate
    network to discover potentially more powerful strategies than those covered by the heuristics. Equally important is the
    automated decision process to choose which behaviors are important. Currently the set of behaviors to be regulated is
    hand-picked based on human players' knowledge and experience about the game. This is obviously domain specific and is difficult 
    to be extended to other games or application domains. 
    Lastly we would like to explore other forms of neural network structure as the current ones are quite basic. 
    More advanced neural networks not only pair well with the ever-strengthening compute, but also may 
    guide the agents to uncover something which we have yet to imagine.

\bibliographystyle{plain}
\bibliography{hp_paper}

\newpage
\appendix
\onecolumn

\section{Playable Combinations}
\label{cardcombos}
\begin{figure}[ht]
    \vskip 0.2in
    \begin{center}
    \centerline{\includegraphics[width=\columnwidth]{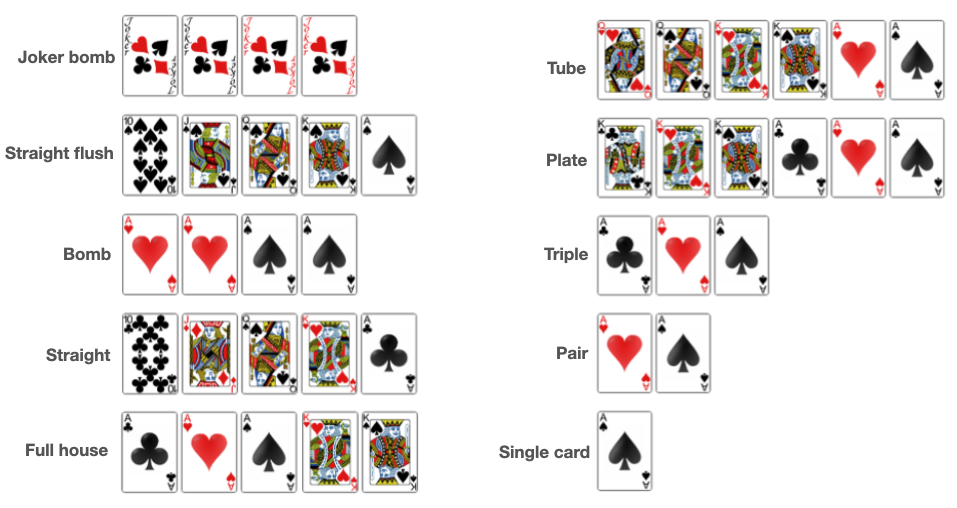}}
    \caption{All playable card combinations in the game of Guandan}
    \label{fig-cardcombos}
    \end{center}
    \vskip -0.2in
\end{figure}

There are many different playable card combinations in the game of Guandan. We describe all the playable combinations below, and also provide
a collection of examples of every combination in \cref{fig-cardcombos}.

\begin{itemize}
    \item \textbf{Single card}: any single card in the two decks being used by the game. The ranks in descending order are: 
    red joker $>$ black joker $>$ level card 
    $>$ A $>$ K $>$ Q $>$ J $>$ 10 $>$ 9 $>$ 8 $>$ 7 $>$ 6 $>$ 5 $>$ 4 $>$ 3 $>$ 2. The rank order also applies to all other types of 
    playable combinations unless specified otherwise.
    \item \textbf{Pair} any two cards which are of the same rank. A pair can be two cards of different suits. However, a black joker 
    and a red joker cannot form a pair.
    \item \textbf{Triple} any three cards which are of the same rank.
    \item \textbf{Plate} two triples whose ranks are consecutive. The rank is determined by the triple of the lower rank. The rank of level cards 
    in a plate is only their face value. In addition, aces can form a plate with three two's, and the rank of the resulting plate is 
    one (the lowest rank of any plate type combinations).
    \item \textbf{Tube} three pairs whose ranks are consecutive. The determination of the rank is similar to that \textbf{plate}, and aces can form
    a tube with a pair of twos and a pair of threes, resulting in a tube of rank one.
    \item \textbf{Full House} a combination of a pair and a triple. The rank of the pair and that of the triple may not be the same. The rank is 
    determined by that of the triple.
    \item \textbf{Straight} five single cards whose ranks are consecutive. The determination of the rank is similar to that of either plate or tube, 
    and once again an ace can form a straight with a two, a three, a four, and a five.
    \item \textbf{Bomb} four to ten cards of the same rank. There are at most eight cards of the same rank in two decks. Two wild cards can help 
    to form a ten-card bomb.
    \item \textbf{Straight flush} a straight with all the cards belonging to the same suit.
    \item \textbf{Joker bomb} two red jokers and two black jokers. This combination is of the highest rank, and can trump all other combinations.
\end{itemize}

When any non-bomb type (single card, pair, triple, plate, tube, full house, and straight) card combination is played, 
the subsequent player must play
a combination of the same type or a bomb. When a bomb is played, the subsequent player must play a bomb of a higher rank, a bomb of a larger card size, or
a straight flush if the previous combination played is either a four-card bomb or a five-card one.

\section{Rules of Cardplay}
\label{rules}
Each mini game starts with the tribute process if the rank of level card of any team is greater than two. 
The first player makes their move once the tribute process finishes. All other players
must choose a legal card combination to play, according to the rules outlined in \cref{cardcombos}. This process continues until 
one of the players empties their hand before all others, and becomes the banker (i.e., the first finished). 
Similarly, the follower (the second finisher) standing can be determined. Immediately after the banker or 
the follower is determined, and no other player can trump the last card combination played by the banker or the follower, the next player must be the 
teammate of the banker or the follower. This is the so called ``the right to play by borrowing the wind''. Once at least three players manage to 
empty their hands, the current mini game concludes. And the winning team upgrades their level card according to the standing of the team members. 
The more dominating the standing is, the greater the level card gets upgraded to. For example, the level card gets upgraded by three levels
if the standing is the best possible combination: ``banker'' and ``follower''. Every possible standing 
combination and the extent of upgrade for the level card is detailed in \cref{tab-level_card}. Each team has their own level card, and there is a 
scoring system (see \cref{tab-score}) to reward/punish the two teams based on the difference between the ranks of the level cards. 
Both teams' level cards are set to two at the start of the first mini game. Multiple mini games are played until a team wins the whole game 
when their level card is ace and no team member attains the dweller standing.

\begin{table}[t]
    \caption{Upgrading the level card based on team members' standing}
    \label{tab-level_card}
    \vskip 0.15in
    \begin{center}
    \begin{small}
    \begin{sc}
    \begin{tabular}{lcr}
    \toprule
    Standing & Upgrade level card by \# \\
    \midrule
    banker, follower & 3 \\
    banker, third & 2 \\
    banker, dweller & 1 \\
    \bottomrule
    \end{tabular}
    \end{sc}
    \end{small}
    \end{center}
    \vskip -0.1in
\end{table}

\begin{table}[t]
    \caption{Scoring system based on the difference between the rank of the level cards}
    \label{tab-score}
    \vskip 0.15in
    \begin{center}
    \begin{small}
    \begin{sc}
    \begin{tabular}{lcccccccccccccccr}
    \toprule
    difference & 0 & 1 & 2 & 3 & 4 & 5 & 6 & 7 & 8 & 9 & 10 & 11 & 12 & 13 & 14 \\
    \midrule
    winning team & 14 & 15 & 16 & 17 & 18 & 19 & 20 & 21 & 22 & 23 & 24 & 25 & 26 & 27 & 28 \\
    losing team & 14 & 13 & 12 & 11 & 10 & 9 & 8 & 7 & 6 & 5 & 4 & 3 & 2 & 1 & 0 \\
    \bottomrule
    \end{tabular}
    \end{sc}
    \end{small}
    \end{center}
    \vskip -0.1in
\end{table}

\section{Tribute}
\label{tribute}
At the start of each mini game, a tribute process must be carried out if a mini game was played previously.
The dweller of the previous mini game must donate their highest-ranked card to the previous banker. In the event of ``double-dweller'', 
where the players on the same team attained the third and the dweller standings, 
both players ought to donate their best cards. And this is called the 
``double tribute'' scenario. The card of the higher rank goes to the banker, the other the follower (the second finisher of a game). 
In the case where the two cards to be donated are of the same rank, the tribute target is then the player sitting farthest to the donating player 
clockwise. 
The tribute process can be denied if both of the red joker cards are owned by the dweller or the double dwellers. If this were to be the case, 
the banker of the previous mini game would lead to play. Otherwise, whoever donates a card to the banker leads to play.

\end{document}